\documentclass{article}

\usepackage{arxiv}

\usepackage[utf8]{inputenc} 
\usepackage[T1]{fontenc}    
\usepackage{hyperref}       
\usepackage{url}            
\usepackage{booktabs}       
\usepackage{amsfonts}       
\usepackage{nicefrac}       
\usepackage{microtype}      
\usepackage{graphicx}
\usepackage{natbib}
\usepackage{doi}
\usepackage{subcaption}
\usepackage{enumitem}
\usepackage{makecell}
\usepackage{tabularx}
\usepackage{multirow}
\usepackage{balance}
\usepackage{xcolor}

\title{Detecting Silent Failures in Multi-Agentic AI Trajectories}

\author{
\textbf{Divya Pathak} \\
IBM Research \\
\texttt{Divya.Pathak2@ibm.com}
\And
\textbf{Harshit Kumar} \\
IBM Research \\
\texttt{harshitk@in.ibm.com}
\And
\textbf{Anuska Roy} \\
IIIT Bangalore \\
\texttt{anuska.roy@iiitb.ac.in}
\And
\textbf{Felix George} \\
IBM Research \\
\texttt{Felix.George@ibm.com}
\And
\textbf{Mudit Verma} \\
IBM Research \\
\texttt{mudiverm@in.ibm.com}
\And
\textbf{Pratibha Moogi} \\
IBM Research \\
\texttt{pratibha.moogi@ibm.com}
}

\date{}

\renewcommand{\headeright}{Preprint}
\renewcommand{\undertitle}{}
\renewcommand{\shorttitle}{Detecting Silent Failures in Multi-Agentic AI Trajectories}

\hypersetup{
pdftitle={Detecting Silent Failures in Multi-Agentic AI Trajectories},
pdfsubject={cs.AI},
pdfauthor={Divya Pathak, Harshit Kumar, Anuska Roy, Felix George, Mudit Verma, Pratibha Moogi},
pdfkeywords={Multi-Agentic AI, Anomaly Detection, Silent Failures, Agentic Trajectories, Non-deterministic Systems},
}

\begin{document}
\maketitle

\begin{abstract}
Multi-Agentic AI systems, powered by large language models (LLMs), are inherently non-deterministic and prone to silent failures such as drift, cycles, and missing details in outputs, which are difficult to detect. We introduce the task of anomaly detection in agentic trajectories to identify these failures and present a dataset curation pipeline that captures user behavior, agent non-determinism, and LLM variation. Using this pipeline, we curate and label two benchmark datasets comprising \textbf{4,275 and 894} trajectories from Multi-Agentic AI systems. Benchmarking anomaly detection methods on these datasets, we show that supervised (XGBoost) and semi-supervised (SVDD) approaches perform comparably, achieving accuracies up to \textbf{98\%} and \textbf{96\%}, respectively. This work provides the first systematic study of anomaly detection in Multi-Agentic AI systems, offering datasets, benchmarks, and insights to guide future research.
\end{abstract}


\section{Introduction}
\label{intro}

The adoption of AI agent systems
has been rapidly increasing across various industries, such as insurance, customer service chatbots, etc~\cite{bhattacharya2025ai, mckinsey2025future, damcogroup2025agents, aimultiple2025chatbots}.
An AI agent consists of a set of tools, large language models (LLMs), and system prompts. These LLMs drive agent decision-making, including selecting tools, invoking other agents, reasoning, and planning, based on system prompts, input queries, context from other agents/tools, and available tools. Due to the reliance on LLMs for decision-making, AI Agents are inherently \textbf{non-deterministic}~\cite{bronsdon2025reliability,cemri2025multi, hammond2025multiagent} i.e, the sequence of execution order of steps (trajectory) in the agent workflow is decided dynamically. 
Moreover, changing the system prompt or the LLM for the same input may result in a different trajectory.
Furthermore, even for the same input query, using the same model and system prompt, the agent’s execution may vary, thereby leading to different trajectories, sometimes erroneous too. 

\begin{table}[!hbtp]
\centering
\caption{Silent Failure Scenarios in Multi-Agentic AI Systems}
\renewcommand{\arraystretch}{1.2}
\begin{tabularx}{0.95\linewidth}{p{5cm} X}  
\toprule
\textbf{Failure} & \textbf{Description} \\
\midrule
Drift & The agent diverges from the intended path, selecting tools or subsequent agents that are irrelevant for an input query. \\ \hline
Cycles & The agent repeatedly invokes itself or other agents/tools by re-planning resulting in redundant loops, wasted computation. \\ \hline
Missing Details in Final Output & The agent returns a response without errors, but misses crucial information requested in the input query. \\ 
\hline
Tool Failures & External tools(APIs) may fail silently, return unexpected results, hit rate limits—issues that the agent may not detect or handle gracefully. \\ \hline
Context Propagation Failures & Failure in propagating the correct context to dependent agents/tools. \\
\bottomrule
\end{tabularx}
\label{silent-failure}
\end{table}

Due to this non-determinism and dynamic nature, Agentic AI systems 
are more prone to failures compared to traditional microservice based applications~\cite{bronsdon2025reliability,cemri2025multi}. 
Unlike microservice applications, where errors are typically explicit (with error codes) and predictable, failures in agentic systems can often be \textbf{silent}, occurring without generating clear error signals while still deviating from the intended behavior. Table~\ref{silent-failure} summarizes the \textit{different failures scenarios} that can arise in such systems. For instance, a drift is when an agent chooses next agent A and tool B instead of agent D and tool E. Such behaviors are observed through Agentic AI traces (Figure~\ref{fig:dataset-curation} $(B)$), which capture the complete execution workflow of an input request across agents, tools, and LLMs. We use the terms trajectories and traces interchangeably throughout this paper. 

Effective detection mechanisms are essential, as such \textit{silent failures} can rapidly escalate operational costs, including \textbf{computational resources, token usage, and time}. We define \textbf{anomaly detection} as the task of detecting the existence of such silent failures in Agentic AI trajectories. Detecting anomalies and evaluating these detection techniques in Agentic AI systems requires datasets that capture diverse types of failures. To the best of our knowledge \textit{no publicly available datasets} currently exist that capture the diverse behaviors and failure scenarios of Multi-Agentic AI systems. Although anomaly detection is extensively explored in areas like microservices and networks~\cite{anomalymicro1,anomalymicro2,networkanomly}, established techniques for AI Agents remain unexplored. Recent work like~\cite{he2025sentinelagent} applies graph-based methods but offers limited evaluation. To bridge this gap, this work makes the following three key contributions: 


\begin{itemize}[leftmargin=1em]
    \item \textbf{Dataset Curation Pipeline (Section~\ref{dataset})}: We design a comprehensive pipeline for curating datasets from agentic traces for the anomaly detection task. This includes collecting traces that capture \textit{user behavior, agent non-determinism, and LLM model variation}, extracting relevant features, labeling traces into two binary classes (normal/anomaly). Using this pipeline, we curate labeled datasets from two Multi-Agentic AI systems, comprising \textbf{4,275} and \textbf{894} datapoints. We plan to open-source these datasets to foster community-driven benchmarking and further research on anomaly detection in agentic systems.
    \item \textbf{Benchmarking Anomaly Detection Methods (Section~\ref{method})}: We leverage the two datasets 
    to benchmark anomaly detection methods, demonstrating their utility for both research and practical applications. Among the evaluated methods, XGBoost~\cite{chen2016xgboost}, a supervised binary classification approach, achieved accuracies of \textbf{98\%} and \textbf{94\%} on the two datasets. While, SVDD~\cite{tax2004support}, a one-class semi-supervised method, obtained \textbf{96\%} and \textbf{89\%}, respectively.
    \item \textbf{Detailed Error Analysis and Insights (Section~\ref{method}):} We conduct a thorough analysis of detection results, identifying causes of misclassifications, areas for improvement, and directions for future research in Agentic AI systems.
\end{itemize}
These contributions offer researchers and practitioners practical guidance for building robust and reliable AI-Agentic systems, and provides insights to current limitations and future directions for developing approaches to detect silent failures.

\section{Dataset Curation Pipeline for Multi-Agentic AI Systems}
\label{dataset}

In this section we present a \textit{dataset curation pipeline} used to curate datasets for anomaly detection task, which can also be extended to other tasks such as drift detection and cycle detection. Figure~\ref{fig:dataset-curation} illustrates the curation pipeline consisting of three components: (1) collect Agentic AI traces,  (2) extract key features from these traces and (3) define labeling criteria to label these traces into two binary classes - normal and anomaly. Finally, we apply this pipeline to construct datasets from two representative Multi-Agentic AI systems: (1) Stock Market Assistant and (2) Research Writing Assistant. This pipeline can be readily extended to other Agentic AI systems. 



\subsection{Agentic Traces: The Building Blocks}
\begin{figure*}[!hbtp]
\centering
    \includegraphics[width=0.99\textwidth]{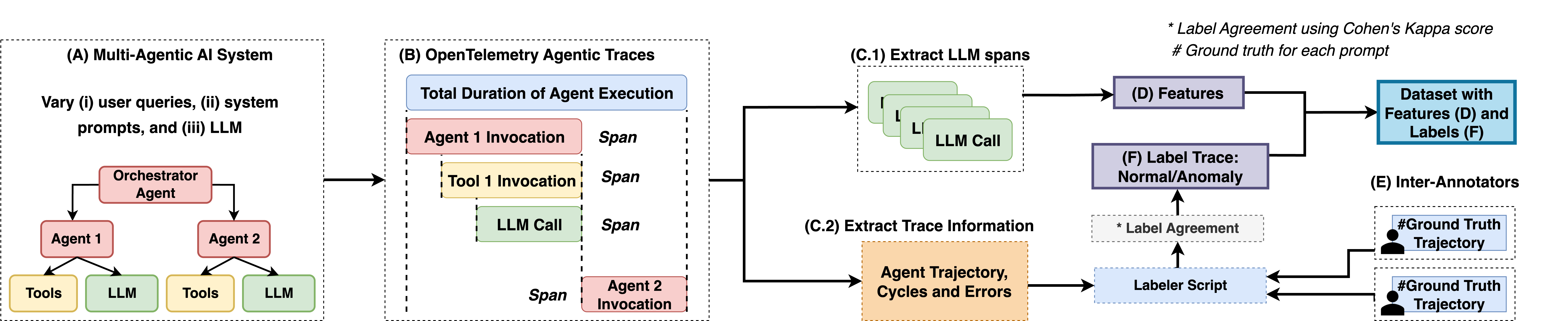}
\caption{Dataset Curation Pipeline for Multi-Agentic AI Systems}
\label{fig:dataset-curation}
\end{figure*}

The first step in our dataset curation process is the collection of agentic traces. Just as traces in microservice based application represent the end-to-end flow of a request across distributed services, Agentic AI traces capture the complete execution workflow of an input request across agents, tools, and LLMs. For instance shown in Figure~\ref{fig:dataset-curation} $(B)$, a request executed by Agent 1 is recorded as a span, which triggers Tool 1 and an LLM call, forming child spans linked to the parent Agent 1 span. To collect agentic traces, we instrument Agentic AI applications using OpenTelemetry distributed tracing~\cite{opentelemetry}, a standard for capturing traces in distributed systems. 


Next, we carefully generate these traces by simulating three key factors that influence agentic behavior: \textbf{Non-determinism of AI agents:} Executing the same or similar query multiple times may result in different trajectories due to variations in reasoning and planning. \textbf{LLM model:} Acts as the decision-making engine of the AI agent, shaping its reasoning and planning. \textbf{User behavior:} Each user or developer has a distinct style of writing system prompts, which influences agent behavior.

To systematically capture these factors in the traces, we vary three key variables: the \textbf{input query prompt}, the \textbf{LLM model employed}, and the \textbf{system prompt}. We categorize system prompts into three types: a \textit{poor} prompt is highly ambiguous and provides minimal structure; a \textit{good} prompt is well-structured and follows the ReAct-prompting pattern~\cite{yao2022react}; and a \textit{strict} prompt extends a good prompt with additional rules and illustrative examples.

\subsection{Feature Extraction from Agentic Traces}
Once traces are collected, the next step is to extract \textbf{\textit{relevant features}} capturing key characteristics of Agentic AI executions. Features are extracted primarily from \textit{LLM call} span attributes, as they drive decision-making, token usage, latency, and reasoning behavior, whereas agent and tool spans contribute minimally to these attributes. 
We extract \textbf{16} features organized into five key categories:
\begin{itemize}[leftmargin=1em]
    \item \textbf{Token Features:} Captures the number of tokens consumed by the LLMs for its input, and output
    as well as aggregated (sum, mean, std) token usage across all the LLM calls in the trace. These feature provides insights into computational cost. 
    \item \textbf{Latency Features:} Including per-span latency, sum and mean latency, and end-to-end total time. These features highlight performance bottlenecks and help benchmark execution efficiency.
    \item \textbf{Path Features:} Representing the sequence of agent, tool, and LLM calls, length of hierarchical delegation, unique agent/tool calls. These features capture the trajectory of the application and extracted using trace structure.
    \item \textbf{Prompt and Context Features:} System and user prompts, prompt lengths, and intermediate inputs/outputs. These features provide semantic context, explaining variations in execution behavior and reasoning patterns.
    \item \textbf{Model Features:} Information about the specific LLM or tool versions used, allowing fair comparison between different models.
\end{itemize}

\subsection{Labeling traces: Normal or Anomalous}
Next is to define a labeling scheme to flag traces as anomalous (experiencing silent failures) or normal (executed successfully). Providing these labels serve as a ground truth and is essential as it enables systematic evaluation, benchmarking, and comparison of anomaly detection task, and help in performance optimization of agentic workflow execution. 
We label a trace as an anomaly if it exhibits any of the three failure scenarios defined in Table~\ref{silent-failure}: \textbf{drift, cycles, or errors}. Drift is detected by domain experts who define the ground truth trajectory for each input prompt, allowing deviations from the expected path to be identified. For cycles, we define a strict criterion: no agent or tool should be invoked more than once within a single execution. Errors are identified directly from the trace spans by checking the error status recorded during execution. 



Next, using a script, we extract trace information such as observed trajectory, cycles, and errors for each trace corresponding to a given input prompt. The inter-annotator defines its own ground truth (expected trajectory) for each input prompt. The inter-annotator uses automatic labeling script (Figure~\ref{fig:dataset-curation}) which takes the ground truth and the extracted information for each trace to automatically assign a label of normal or anomalous. This labeler script is highly extensible, as future annotators only need to provide the ground truth for new prompts, and the script can automatically generate the corresponding labels.

We apply the pipeline\footnote{The dataset and curation pipeline will be released after paper acceptance in accordance with organizational policies.} (Figure~\ref{fig:dataset-curation}) to \textbf{two} Multi-Agent AI systems:
\begin{itemize}[leftmargin=1em]
    \item \textbf{Stock Market Analysis Assistant:} This system has 3 agents and 9 tools -- an orchestrator overseeing \textit{stock\allowbreak\_market} and \textit{search\allowbreak\_agent}, calling tools such as \textit{stock\allowbreak\_quote, stock\allowbreak\_recommendation, company\allowbreak\_transactions, search\allowbreak\_symbol, timeseries\allowbreak\_daily\allowbreak\_info, search\allowbreak\_internet.} Traces were generated using 525 user prompts, 3 system prompt types (poor, good, strict), and 3 LLMs (gpt-4o, ibm-granite-3-1-8B, meta-llama-3-3-70B), producing 4,275 datapoints. Two annotators labeled traces (Cohen’s kappa~\cite{cohen1960coefficient} 97.6\%). Among the datapoints, 1,804 are normal and 2,921 anomalous, including cycles (1,484), errors (1,245), and drift (1,952), note that these categories are non-exclusive.
    \item \textbf{Research and Writing Assistant:} This system has 9 agents and 6 tools -- an orchestrator overseeing research and writing orchestrator agents, which call multiple helper agents (\textit{doc\allowbreak\_writer, note\_taker, chart\allowbreak\_maker, rag\allowbreak\_agent, web\allowbreak\_scraper, search\allowbreak\_internet} each with one tool). Traces were generated from 112 user prompts, 2 system prompt types (good, poor), and 2 LLM models (ibm-granite-3-1-8B, meta-llama-3-3-70B/405B), producing 894 datapoints. Two annotators labeled the traces (Cohen’s kappa 80.6\%). Of these, 321 are normal and 573 are anomalous, including cycles (320), errors (285), and drift (410). 
\end{itemize}

\section{Anomaly Detection in Multi-Agentic AI Systems}
\label{method}


In this section, we leverage the curated datasets to perform anomaly detection task. We start by understanding the datasets and visualizing feature distributions, then describe the experimental setup in both supervised, semi-supervised, and unsupervised settings. Finally, we present the results, highlighting feature importance, error analysis, and key insights.

\subsection{Dataset Feature Space Visualization}
\begin{figure}[h]
    \centering
    \begin{subfigure}[b]{0.23\textwidth}
        \includegraphics[width=\textwidth]{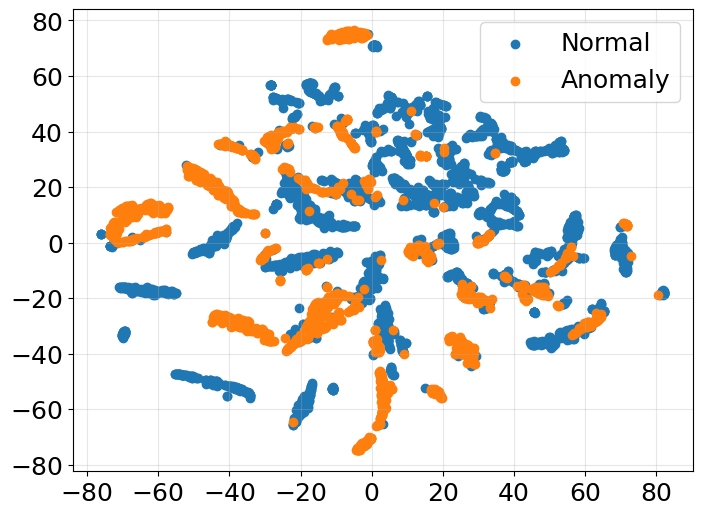}  
        \caption{Stock Market Dataset}
        \label{fig:tsne_stock}
    \end{subfigure}
    \begin{subfigure}[b]{0.23\textwidth}
        \includegraphics[width=\textwidth]{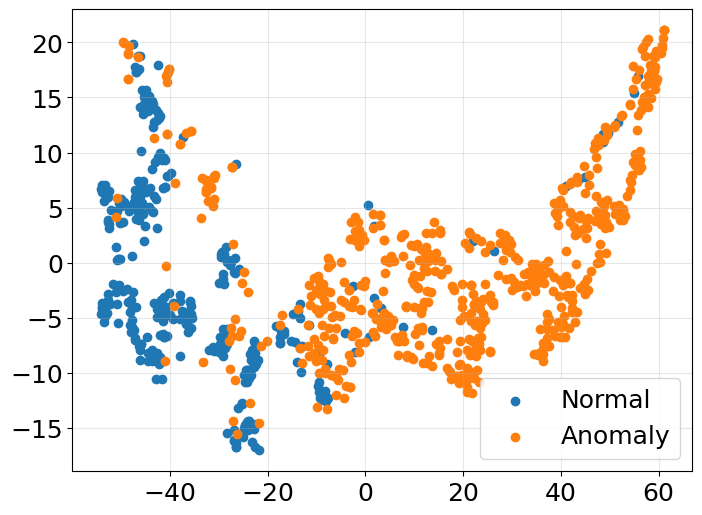}  
        \caption{Research Writing Dataset}
        \label{fig:tsne_research}
    \end{subfigure}
    \caption{t-SNE of normal and anomalous agentic traces}
    \label{fig:tsne}
\end{figure}

To understand the feature space, we project the 16-dimensional features using t-SNE~\cite{tsne2008}. Figure~\ref{fig:tsne} shows the 2D distribution of normal and anomalous traces. In the stock market (Figure~\ref{fig:tsne_stock}), some anomalous points overlap with normal, indicating higher detection complexity, whereas the research writing (Figure~\ref{fig:tsne_research}) shows clearer separation. This motivates the need for sophisticated models as feature distributions vary across applications.

\begin{table*}[h]
\centering
\caption{Comparison of Model Performance Across Two Datasets. $^{*}$For Anomaly (Positive) Class}
\label{tab:results}
\begin{tabular}{p{1.5cm} p{1.5cm} p{1.2cm} p{1.2cm} p{1.2cm} p{1.2cm} p{1.2cm} p{1.2cm} p{1.2cm} p{1.2cm}}
\toprule
\textbf{Method Type} & \textbf{Model} & \multicolumn{4}{c}{\textbf{Stock Market Dataset}} & \multicolumn{4}{c}{\textbf{Research Writing Dataset}} \\
\cmidrule(lr){3-6} \cmidrule(lr){7-10}
 & & Accuracy & Macro-F1 & Precision$^{*}$ & Recall$^{*}$ & Accuracy & Macro-F1 & Precision$^{*}$ & Recall$^{*}$ \\
\midrule
Supervised & XGBoost & \textbf{0.9803} & \textbf{0.9793} & \textbf{0.9977} & 0.9703 & \textbf{0.9481} & \textbf{0.9426} & 0.9341 & \textbf{0.9884} \\
 & Random Forest & 0.9704 & 0.9690 & 0.9952 & 0.9566 & 0.9407 & 0.9347 & 0.9333 & 0.9767 \\
 & Logistic Regression & 0.9464 & 0.9438 & 0.9717 & 0.9406 & 0.9185 & 0.9115 & 0.9310 & 0.9419 \\
 & SVM & 0.9732 & 0.9717 & 0.9816 & \textbf{0.9749} & 0.9111 & 0.9011 & 0.9022 & 0.9651 \\
 & Naïve Bayes & 0.8815 & 0.8789 & 0.9707 & 0.8333 & 0.8444 & 0.8402 & \textbf{0.9577} & 0.7907 \\
\midrule
Semi-Supervised \\ (OCC) & SVDD & 0.9647 & 0.9628 & 0.9791 & 0.9635 & 0.8963 & 0.8820 & 0.8750 & 0.9767 \\
 & Isolation Forest & 0.8914 & 0.8803 & 0.8706 & 0.9680 & 0.8889 & 0.8793 & 0.9080 & 0.9186 \\
\midrule
Unsupervised & KMeans & 0.8533 & 0.8359 & 0.8327 & 0.9543 & 0.8296 & 0.8220 & 0.9091 & 0.8140 \\
\bottomrule
\label{results}
\end{tabular}
\end{table*}


\subsection{Experimental Setup}
To perform anomaly detection task, the datasets for the two applications — The Stock Market dataset (4,275 traces; 1,804 normal, 2,921 anomalous) and Research Writing dataset (894 traces; 321 normal, 573 anomalous) was split into 70-15-15\% for train-validation-test. We consider anomaly as \textbf{positive class}, normal as \textbf{negative class}. We employ three settings enabling a comprehensive evaluation of accuracy, robustness, and generalizability: \textbf{(A) Supervised:} Includes classical ML models like Tree-based models (XGBoost, Random Forest), linear models (Logistic Regression, Gaussian Naïve Bayes), and SVMs. 
These serve as upper-bound baselines assuming both normal and anomalous labels are available.
\textbf{(B) Semi-Supervised:} Given the relative ease of collecting normal traces, we include one-class classifiers (OCC) such as SVDD and Isolation Forest which model the normal behavior and flag deviations as anomalies.
\textbf{(C) Unsupervised:} Since labeling datasets is expensive and not always feasible, we use K-Means clustering to flag anomalies as points in small or low-density clusters. Hyperparameters were tuned via grid search on the validation set. 

\subsection{Results and Discussion}

Table~\ref{tab:results} summarizes the performance of supervised, semi-supervised, and unsupervised models on the Stock Market and Research Writing Assistant datasets. On the Stock Market, \textbf{XGBoost} achieves the best results overall, with accuracy of \textbf{98.03\%}, macro-F1 of \textbf{97.93\%}, and precision of \textbf{99.77\%}. Recall (\textbf{97.03\%}) is slightly lower (\textbf{-0.46\%}) than SVM (\textbf{97.49\%}). On the Research Writing, \textbf{XGBoost} again outperforms other models with accuracy of \textbf{94.81\%}, macro-F1 of \textbf{94.26\%}, and recall of \textbf{98.84\%}. Naive Bayes achieves the highest precision (\textbf{95.77\%}, \textbf{+2.36\%} over XGBoost). Semi-supervised \textbf{SVDD} also perform competitively (up to \textbf{96.47\%} accuracy on Stock Market and \textbf{89.63\%} on Research Writing), showing their ability to detect anomalies even when trained only on normal traces. On the Research Writing, Isolation Forest achieves higher precision than SVDD (\textbf{+3.30\%}). Compared to supervised methods, unsupervised \textbf{K-Means} yields moderate performance across both datasets, highlighting the challenges of anomaly detection in the absence of labeled data.

Overall, \textbf{XGBoost} emerges as the best-performing supervised model, while in the semi-supervised setting, the one-class approach \textbf{SVDD} performs the best. However, there remains significant room for improvement in the unsupervised case. Notably, XGBoost and SVDD outperform inter-annotator agreement—97.6\% for Stock Market and 80.6\% for Research Writing—indicating
misclassification are likely due to ambiguous traces where even humans disagree. This highlights that while supervised and semi-supervised models perform comparable, the cost and difficulty of annotation make semi-supervised methods more practical for real-world applications. Next, we analyze \textbf{feature importance} for models (XBoost and SVDD) — using SHAP~\cite{shap2017} to identify which features most strongly influence anomaly detection. There are several common features that appear in the \textbf{top-10} for both models across the two datasets. \textbf{Path-level features} such as tool\_count, total\_steps, unique\_steps, and agent\_count—are consistently \textbf{ranked highest}, highlighting their critical role in anomaly detection. While some \textbf{latency and prompt features} such as total\_duration, std and avg\_prompt\_similarity features also appear, they tend to have \textbf{lower rankings} in terms of feature importance.

To understand model misclassifications, we conducted a detailed \textbf{error analysis}. For each false negative (FN, anomaly predicted as normal), we compared the mean feature values to those of the corresponding ground-truth class (TPs) to understand which features most strongly contributed to misclassification. For the Stock Market dataset, XGBoost has \textit{13} FNs and SVDD \textit{16} out of \textit{438} anomalies, with all \textit{13} XGBoost FNs overlapping with SVDD. For Research Writing, XGBoost has \textit{1} FN and SVDD \textit{2} out of \textit{86} anomalies, with \textit{1} XGBoost FN overlapping with SVDD. While FPs also occur, FNs are key to understand model weaknesses in our use case. Upon analyzing the FNs, several interesting insights emerge: \textbf{(Insight 1)} Most anomalies (TP) involve cycles and errors, which also caused drift and increase path-level feature values 
making the models easier to predict them as anomalies. \textbf{(Insight 2)} Some anomalies follow shorter, drifted paths without explicit cycles or errors, having very less value for path-level, prompt \& latency features \textit{(negative difference with TP mean)} closely resembles normal traces and thus harder to flag as anomaly leading to misclassifications.  
\textbf{(Insight 3)} t-SNE (Figure~\ref{fig:tsne}) support these findings: anomalies with cycles/errors form well-separated clusters, while anomalies with subtle drift overlap with normal traces. To conclude, these FNs though few in number are strong examples of silent failures where the system drifts subtly while feature values remain within normal ranges flagging them as normal by both XGBoost and SVDD, a key area for improvement in future works.

\section{Conclusions and Future Plans}

In this work, we introduced anomaly detection of agentic traces in Multi-Agentic AI systems, addressing challenges like non-deterministic behavior and silent failures. We curated datasets from traces of two such systems and evaluated on supervised, semi, unsupervised methods. Our results show that while XGBoost and SVDD perform comparable in both supervised and semi-supervised settings, preparing labeled data is easier in the semi-supervised setting. Error analysis indicates that a small number of false negatives persist, where both models miss anomalies due to their close resemblance to normal behavior. As part of \textbf{future work}, we plan to enhance anomaly detection methods by thoroughly analyzing feature statistics, particularly focusing on anomalies that mimics normal behavior. This includes investigating additional feature engineering strategies beyond the current set to improve model sensitivity. Given the high cost and scarcity of labeled data, developing robust unsupervised or semi-supervised approaches suitable for both offline and online settings will also be a key focus. Further analysis is needed to explore and address the other failure modes identified in our study (see Table~\ref{silent-failure}), aiming to expand the coverage and robustness of detection methods across varied silent failure scenarios.

\bibliographystyle{unsrtnat}
\bibliography{sample-base}

\end{document}